\documentclass[10pt]{article} % For LaTeX2e

\usepackage[accepted]{rlvg} % Should be uncommented for the camera-ready
\usepackage{amssymb}            % Defines common symbols like \mathbb R
\usepackage{mathtools}          % Extends amsmath, providing common math tools
\usepackage{mathrsfs}           % Enables \mathscr, which can work in cases that \mathcal does not
\usepackage{graphicx}           % For including images
\usepackage{subcaption}         % Allows for the use of subfigures and subcaptions
\usepackage[space]{grffile}     % For spaces in image names
\usepackage{url}                % For displaying URLs
\usepackage{lipsum}             % For placeholder text

\usepackage{tikz}
  \usetikzlibrary{positioning, arrows.meta, shapes.geometric}
\usetikzlibrary{trees, fit,backgrounds, calc, fadings}
\usepackage{svg}
\usepackage{amsmath}
\usepackage{dsfont}
\usepackage{booktabs}

\title{FootsiesGym: A Fighting Game Benchmark for Two-Player Zero-Sum Imperfect-Information Games}
\setrunningtitle{FootsiesGym: A Fighting Game Benchmark for Two-Player Zero-Sum Imperfect-Information Games}
\author{
Chase McDonald\textsuperscript{1},
Nathan Tsang\textsuperscript{2},
Wesley N. Kerr\textsuperscript{2}
}

\emails{chase@comoresearch.org, \{ntsang, wkerr\}@riotgames.com}

\affiliations{
$^{1}$\textbf{Como Research}\\
$^{2}$\textbf{Riot Games}\\

}

\begin{document}

\maketitle  % Make the title section

\begin{abstract}
We present \textbf{FootsiesGym}, an open-source environment for learning in a non-trivial two-player, zero-sum, imperfect-information game. Built on HiFight's minimalist 2D fighting game \emph{Footsies}, it isolates the cyclic, non-transitive strategic interactions of fighting game neutral play while remaining simple enough for efficient analysis. We provide a vectorized simulator that enables high-throughput training on standard hardware, making the environment accessible and reproducible. We describe the design of the environment, benchmark several reinforcement learning algorithms, and discuss open research directions it enables. The code is available at \url{https://github.com/como-research/FootsiesGym}.
\end{abstract}

%%%%%%%%%%%%%%%%%%%%%%%%%%%%%%%%%%%%%%%%%%%%%%%%%%%%%%%%%%%%%%%%
%% Section: Submission of papers to RLVG
%%%%%%%%%%%%%%%%%%%%%%%%%%%%%%%%%%%%%%%%%%%%%%%%%%%%%%%%%%%%%%%%
\section{Introduction} \label{sec:introduction}

% =====================================================================
% Page budget: ~0.5 page.
% Goal: motivate the gap, name the contributions, point at the system fig.
% =====================================================================

Progress in multi-agent and game-theoretic reinforcement learning depends on benchmarks that 
capture the right structure at a tractable cost. Existing environments tend toward two extremes. 
On one side are clean game-theoretic benchmarks such as matrix games and poker variants, which expose cyclic, mixed-strategy structure but typically have short horizons, simple dynamics, and limited exploration requirements. On the other are large, real-time environments (e.g., StarCraft II and Dota 2), which are deeply complex with long time horizons, but require extensive resources for training.

Fighting games can occupy a productive middle ground: they are real-time and spatial, yet keep the strategic structure front and center. They rely on fundamentally 
non-transitive dynamics based on the interaction between two opposing players. The basic strategy
in fighting games is often described as a game of \emph{rock-paper-scissors}, as players must learn
mixed strategies over the primitives of movement and the different kinds of attacks that counter each other and form strategy cycles. Because no single option dominates, a player who commits to a pure strategy can be easily exploited.

\begin{figure}[ht]
  \centering
  \includegraphics[width=0.5\textwidth]{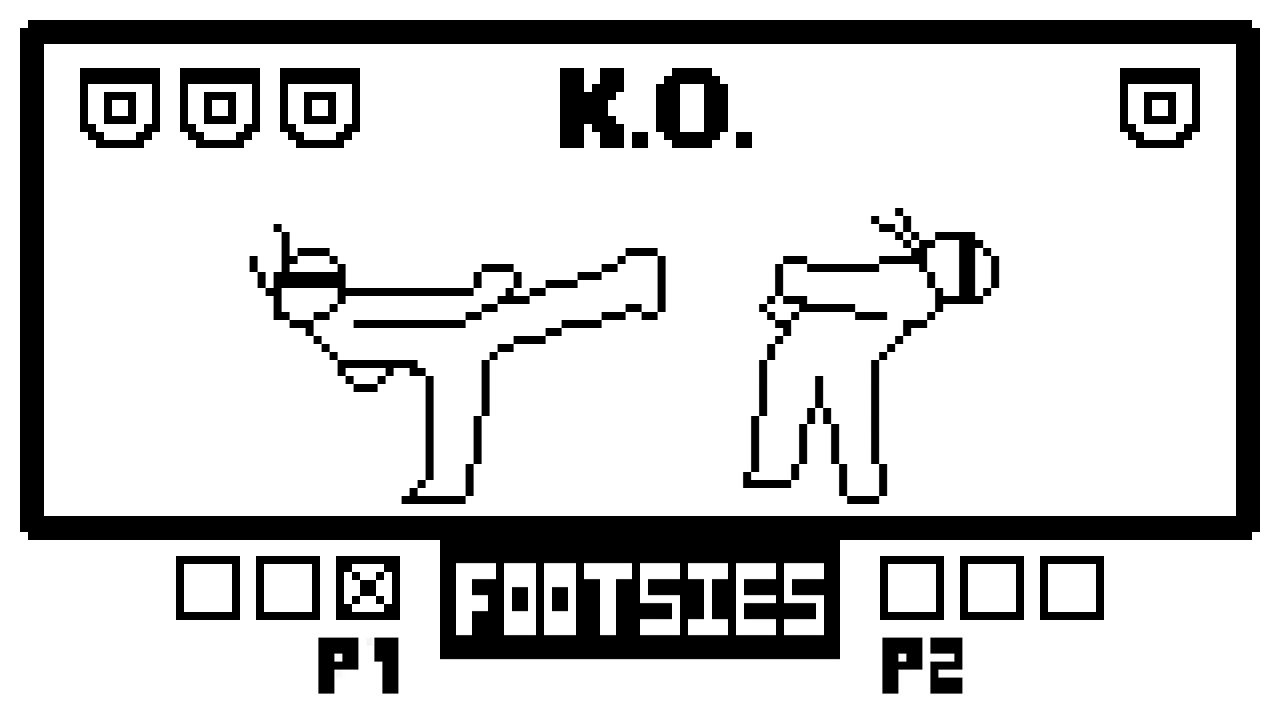}
  \caption{A screenshot of the Footsies game, from \citet{hifight2018footsies}. Two characters fight one another using a combination of movement, quick attacks, and special attacks. A player wins when they K.O. the other.}
  \label{fig:footsies_screenshot}
\end{figure}

In this work, we provide a benchmark environment to study these interactions in Footsies~\citep{hifight2018footsies}, an open-source 2D fighter by HiFight, shown in Figure~\ref{fig:footsies_screenshot}. Footsies is built around the \emph{neutral game}: the phase in which neither player has a clear advantage, and both players maneuver, adjust their spacing and timing, and attempt to create an opening to attack. This is precisely where the cyclic, mixed-strategy structure described above emerges.

\section{Related Work} \label{sec:related_work}

\paragraph{Benchmark Environments.}
Multi-agent and game-theoretic reinforcement learning research spans a wide range of
environment complexity. On one end, there are the small-scale environments that permit exact exploitability calculations~\citep{lanctot2019openspiel,rudolph2025reevaluating}, such as Kuhn Poker, Goofspiel, Leduc Poker, Phantom Tic-Tac-Toe, and Dark Hex. While these environments are valuable for exact exploitability calculations and equilibrium analysis, they capture little of the complexity of richer games. On the other end, real-time strategy and MOBA environments such as StarCraft~II~\citep{vinyals2019grandmaster} and Honor of Kings~\citep{ye2020mastering} have rich, long-horizon dynamics but demand substantial compute and time to train. Between these extremes, environments such as Slime Volleyball~\citep{evojax2022} and simplified Stratego variants~\citep[e.g.,][]{mcaleer2020pipeline} offer intermediate complexity at moderate cost. FootsiesGym likewise occupies this middle ground: it is real-time, spatial, imperfect-information, and non-transitive, yet remains cheap enough to train on a single workstation in a reasonable amount of time.

\paragraph{Fighting games as RL environments.}
There are several instances of fighting games being used as RL environments. The FightingICE
platform~\citep{lu2013fighting, khan2022darefightingice} has hosted a long-running fighting-game AI competition,
while DIAMBRA Arena~\citep{palmas2022diambra} and Stable-Retro~\citep{farama2023stableretro} expose classic arcade and console fighting games through emulation. These existing platforms typically wrap proprietary games that require user-supplied ROMs or closed-source engines, adding licensing and reproducibility friction. FootsiesGym instead builds on a fully open-source game with a purpose-built, vectorized simulator, prioritizing accessibility and training speed on standard hardware.
 
\section{FootsiesGym Environment} \label{sec:env}

% =====================================================================
% Page budget: ~0.8 page total across 3.1-3.4.
% RL-env description, not a systems post-mortem. Keep each subsection ~4-6
% lines of prose unless it owns a figure or table.
% =====================================================================

Compared to existing fighting game environments, Footsies is intentionally minimalist, allowing for easier analysis and a focus on the genre's non-transitive structure. Typical fighting games feature more complex moves and mechanics, most notably combos: long sequences of chained attacks. These mechanics separate fighting games into two strategic regimes: the cyclic, mixed-strategy interaction that arises during the neutral game, and the largely transitive contest of execution that begins once a player lands an opening and starts a combo. By foregoing the latter, Footsies isolates the neutral game while preserving its cyclic strategic interactions.  The key mechanics, restated from the Footsies documentation,\footnote{\url{https://github.com/hifight/Footsies}} are:
\begin{itemize}
    \item There is no health bar; the round is lost when a player is hit by a special attack.
    \item Opponent attacks can be blocked up to three times, indicated by the guard bar on the top right and left of the screen. After three blocks, every attack causes a guard break that leaves the player open to a follow-up attack.
    \item Players can dash forward and backward by pressing the direction button twice in quick succession. 
    \item Special attacks are performed by holding the attack button for 60 frames and then releasing it, from neutral or with a forward/backward direction.
\end{itemize}

\paragraph{Actions.}
Each agent's per-step action is drawn from a discrete space of size
$|\mathcal{A}|=6$ by default, comprising
\textsc{none}, \textsc{back}, \textsc{forward}, \textsc{attack},
\textsc{back+attack}, and \textsc{forward+attack}. Special attacks in Footsies are executed by holding the attack button for a fixed duration and then releasing it. Because discovering when to begin charging and how long to hold the attack button through random exploration can be difficult, FootsiesGym optionally provides a special-charge action mode. When enabled, the action space expands to $|\mathcal{A}|=9$ by adding three charge actions: \textsc{back-charge}, \textsc{neutral-charge}, and \textsc{forward-charge}. Each charge action toggles the held-attack ``charge'' state required to fire a special move: the first invocation begins holding the attack button, while the second releases it and executes the corresponding special move (back, neutral, or forward). Appendix~\ref{sec:appendix_sp_charge} provides further details and illustrates the effect of enabling this option.

\paragraph{Observations.}
Each agent receives an $85$-dimensional featurized observation
($88$-dimensional with the special-charge option enabled)
comprising the features required to fully describe the state of both players.
Each agent additionally observes a small block of privileged self-features, which includes its
own dash availability, special-attack charging progress, and previous action. These
are not exposed in the opponent block, mirroring the
information asymmetry present in human play. The
complete per-feature breakdown is given in
Appendix~\ref{sec:appendix_obs}.

\paragraph{Rewards.}
Rewards are zero-sum between agents ($r^i = -r^{-i}$), where $i$ represents self, and $-i$ represents the opponent. A terminal
win/loss reward of $\pm w$ (default $w=10$) is issued when either agent is defeated. In the event of a timeout (default 4,000 steps), both agents receive 0 reward. Optional dense rewards can be set in the environment configuration to guide learning (see Appendix~\ref{sec:appendix_config}).

\subsection{Action Delay} \label{sec:hyperparameters}

Action delay is the number of game frames
between action selection and in-game execution. It must be an integer
multiple of the frame skip, such that an action chosen at step $t$ is
executed at step
$t + \frac{\text{action delay}}{\text{frame skip}}$. The delay
determines which attacks are \emph{reactable} versus which must be
\emph{predicted}. At an action delay of 0, the policy has frame-perfect reactions. In this setting, we observe that gameplay collapses into degenerate, purely reactive strategies. As the delay increases, moves become increasingly unreactable, reducing the neutral game to pure prediction over the opponent's intent and, at high values of action delay, producing near-random play. We provide additional results on the effect of manipulating action delay in Appendix~\ref{sec:appendix_delay}. 

\subsection{Implementation} \label{sec:architecture}

Footsies is built in the Unity game engine, with all game logic written in C\#. For training, we decouple the simulator from Unity's rendering loop and run it headlessly. Vectorization is handled entirely on the C\# side: a single headless process steps $K$ independent game instances in parallel and exposes them behind one gRPC port. Python clients instantiate the environment following the PettingZoo~\citep{terry2021pettingzoo} API. The number of in-process environments $K$ is passed in the config. Standard headless-mode optimizations (e.g., uncapped frame rate,
disabled audio, no rendering) are applied automatically, and each vectorized instance is auto-reset on termination.

\begin{figure}[t]
  \centering
  \includegraphics[width=0.5\linewidth]{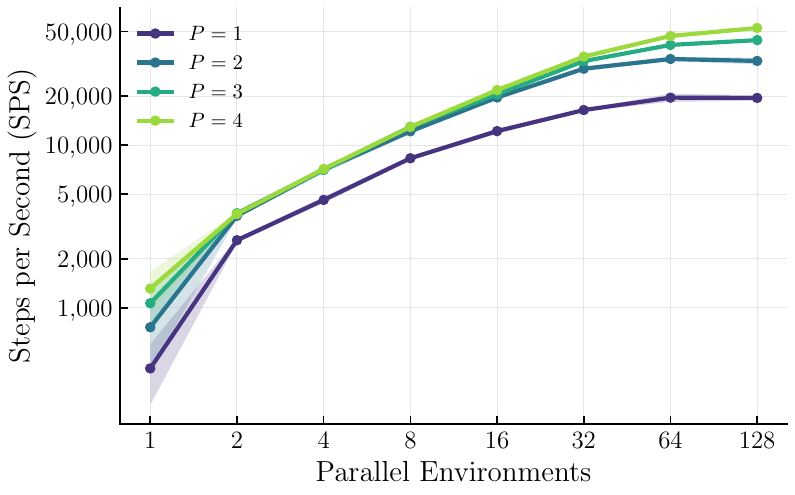}
  \caption{Aggregate environment steps per second as a function of the
    number of in-process parallel environments $K$, on a 24-core / 48-thread workstation. Each curve corresponds to a different number of independent game-server processes $P$ stepped concurrently from Python threads. Shaded band is the 95\% confidence interval.}
  \label{fig:throughput_scaling}
\end{figure}

Figure~\ref{fig:throughput_scaling} characterizes two complementary
layers of parallelism. Within a single game process ($P{=}1$),
aggregate throughput plateaus at roughly $19{,}000$ environment
steps per second. Running $P$ independent game
processes in concurrent threads, each with its own gRPC port and $K$ in-process instances, raises this ceiling. At $P{=}4$ and $K{=}128$, aggregate throughput reaches roughly $52{,}500$ environment steps per second, roughly $2.7\times$ the single-process peak.

In addition to the headless training server, the same Unity project
compiles to a windowed build with full rendering, used for visual
inspection of trained agents and local play from a keyboard.

\section{Experimental Baselines} \label{sec:experiments}

We train a set of baseline algorithms to illustrate the learning dynamics in Footsies and highlight opportunities for future research. Importantly, the reported algorithms are only na\"{\i}vely tuned through several trial-and-error iterations; they are not fully tuned implementations of these algorithms, and we make no definitive claim about their relative quality.

\paragraph{Algorithms.}
We use two variations of Proximal Policy Optimization~\citep[PPO;][]{schulman2017proximal}: one with a large entropy coefficient that anneals over the course of training and another with a fixed coefficient. We refer to these as PPO (Sched.) and PPO, respectively. This selection was motivated by the results of~\citet{rudolph2025reevaluating}, who illustrated that appropriately tuned entropy schedules in standard policy gradient methods can perform comparably or better than specialized game-theoretic training algorithms in cyclic two-player zero-sum imperfect-information games.

\begin{figure}[h]
  \centering
  \begin{subfigure}[t]{0.48\linewidth}
    \centering
    \includegraphics[width=\linewidth]{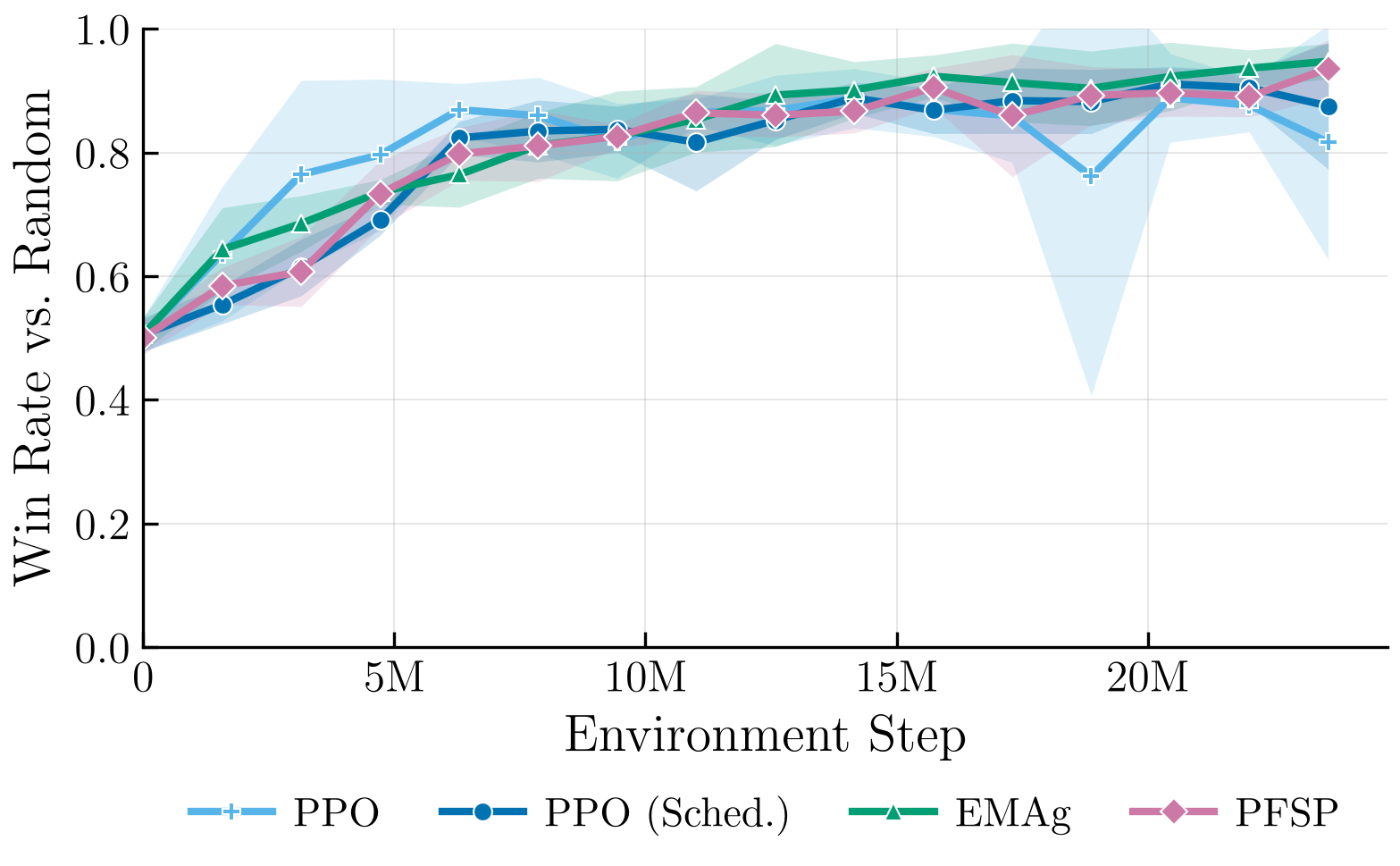}
    \caption{vs.\ random opponent.}
    \label{fig:ppo_curve_random}
  \end{subfigure}
  \hfill
  \begin{subfigure}[t]{0.48\linewidth}
    \centering
    \includegraphics[width=\linewidth]{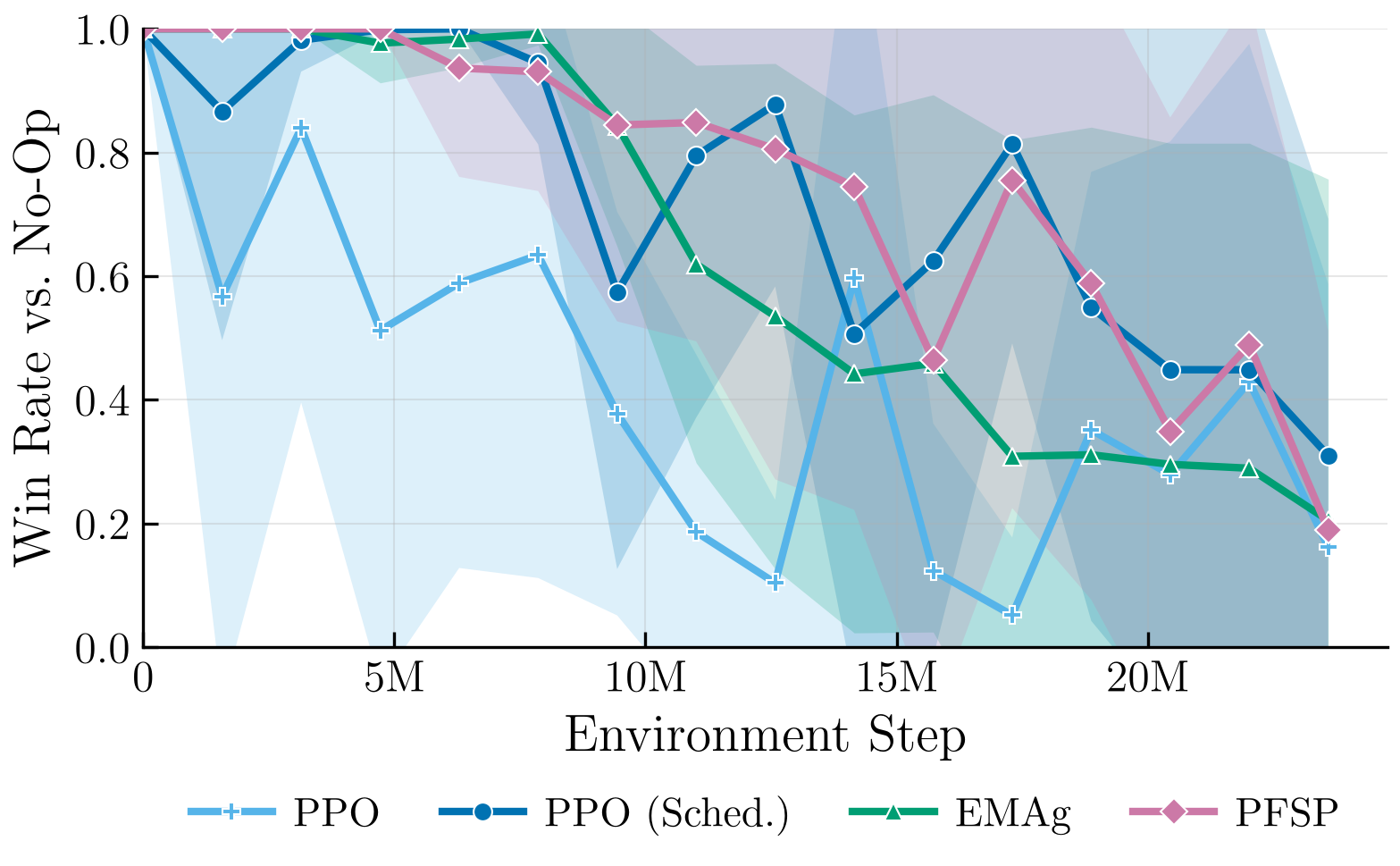}
    \caption{vs.\ no-op opponent.}
    \label{fig:ppo_curve_noop}
  \end{subfigure}
  \caption{PPO, PPO (Sched.), EMAgnet, and PFSP baselines evaluated against
    two heuristic opponents: a (\subref{fig:ppo_curve_random})~uniform-random policy
     and a (\subref{fig:ppo_curve_noop})~no-op policy. Curves are aggregated over 5 seeds;
    the shaded region represents 95\% confidence intervals.}
  \label{fig:winrate_curves}
\end{figure}

In addition, we illustrate training with EMAgnet~\citep{maidment2026emagnet} and a population-based approach using the sampling rule of Prioritized Fictitious Self-Play~\citep[PFSP;][]{vinyals2019grandmaster}. The former is a regularization method for policy gradient algorithms that regularizes the policy towards an exponential moving average of its own weights, and it has been shown to outperform entropy regularization in standard benchmarks. The latter maintains a reservoir of past checkpoints and preferentially samples opponents that the current policy loses to. Full details on both algorithms are provided in Appendix~\ref{sec:appendix_alg_details}.

For all algorithms, we use an identical network structure and training regime: a 2-layer MLP over the 85-dimensional observation space. Each algorithm is trained for $2.5\times 10^7$ environment steps with five random seeds. The complete set of parameters and environment configuration settings is provided in Appendix~\ref{sec:appendix_alg_details}.

\paragraph{Performance Against Heuristic Opponents.}
In Figure~\ref{fig:winrate_curves}, we show the win rate against uniform random and no-op opponents over the course of training. All policies achieve comparable performance against both opponents: their win rate increases to roughly 85--95\% against the uniform random opponent. Because the no-op opponent never attacks, all non-winning matches end in timeouts, so the tie rate is simply $1 - p_{\mathrm{win}}$.  The results illustrate one of many areas for investigation that can be addressed in this environment. As the policies improve, they become increasingly reactive and less likely to initiate an engagement with an opponent who stands still. Although this behavior may be less exploitable, it may be undesirable. For example, it may go against the expectations of a game designer if their in-game AI fails to engage with players.

\begin{figure}[h]
  \centering
    \begin{subfigure}[t]{0.48\linewidth}
    \centering
    \includegraphics[width=\linewidth]{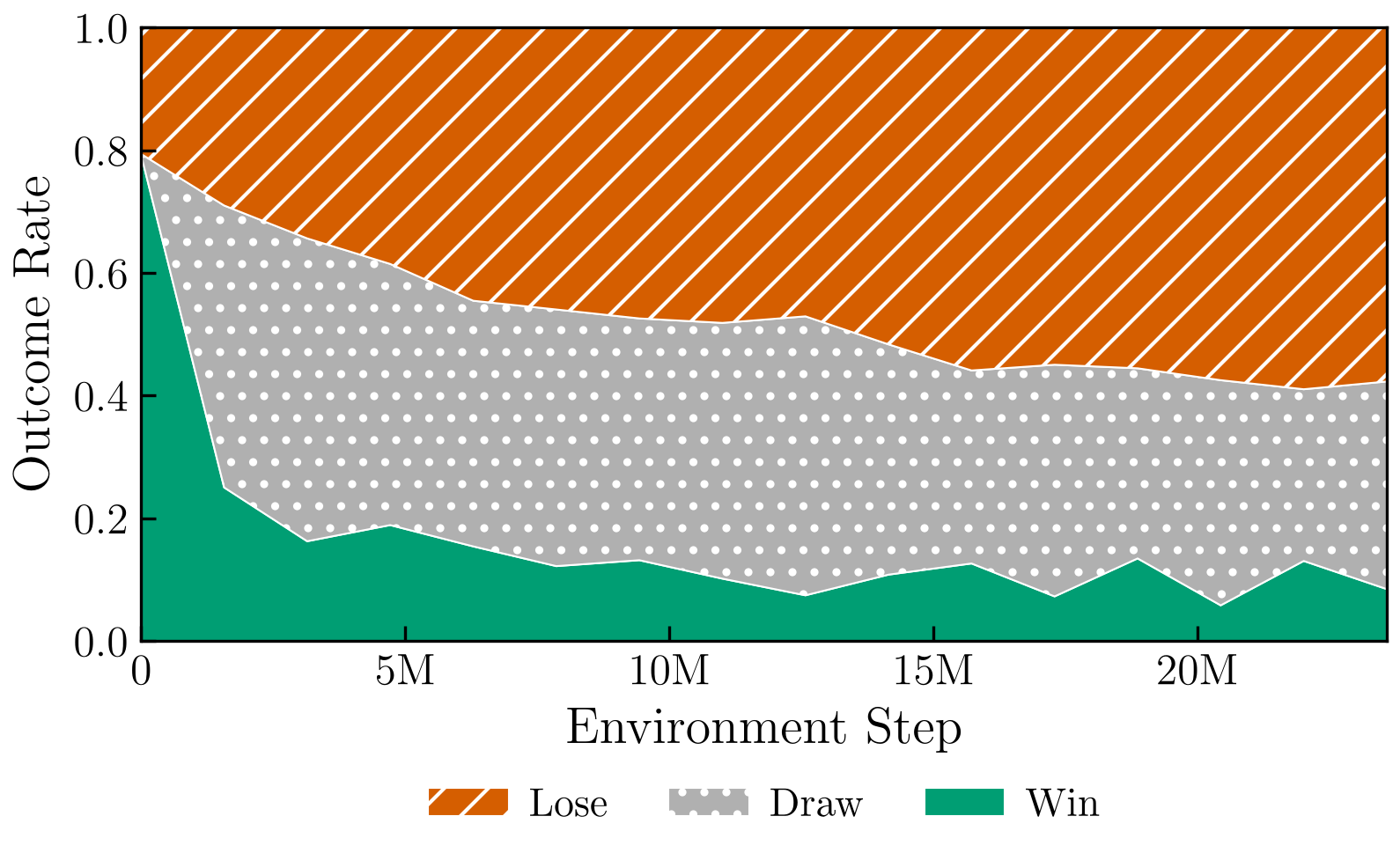}
    \caption{PPO}
    \label{fig:ppo_exploiter}
  \end{subfigure} 
  \begin{subfigure}[t]{0.48\linewidth}
    \centering
    \includegraphics[width=\linewidth]{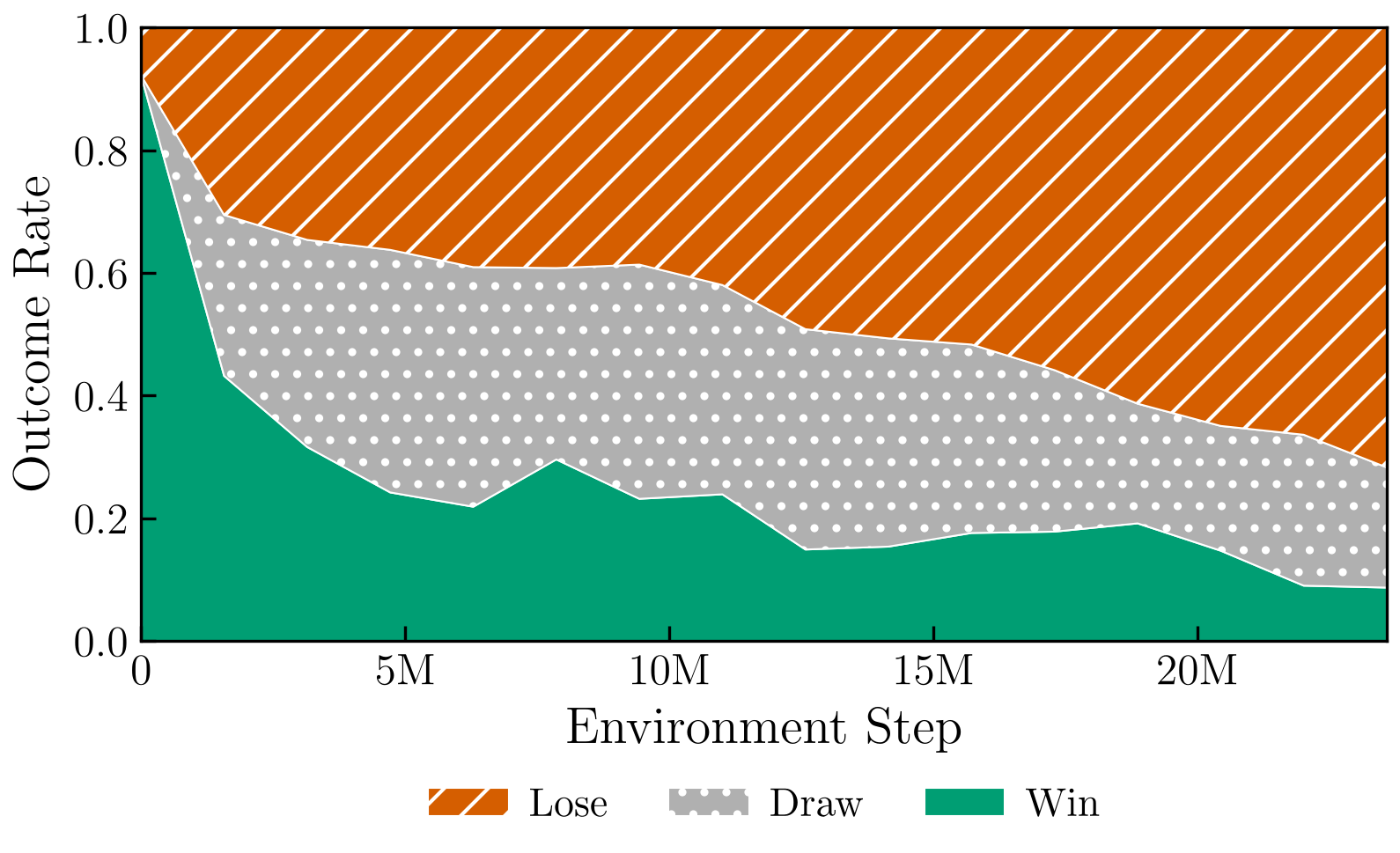}
    \caption{PPO (Sched.)}
    \label{fig:ppo_sched_exploiter}
  \end{subfigure} \\
  \begin{subfigure}[t]{0.48\linewidth}
    \centering
    \includegraphics[width=\linewidth]{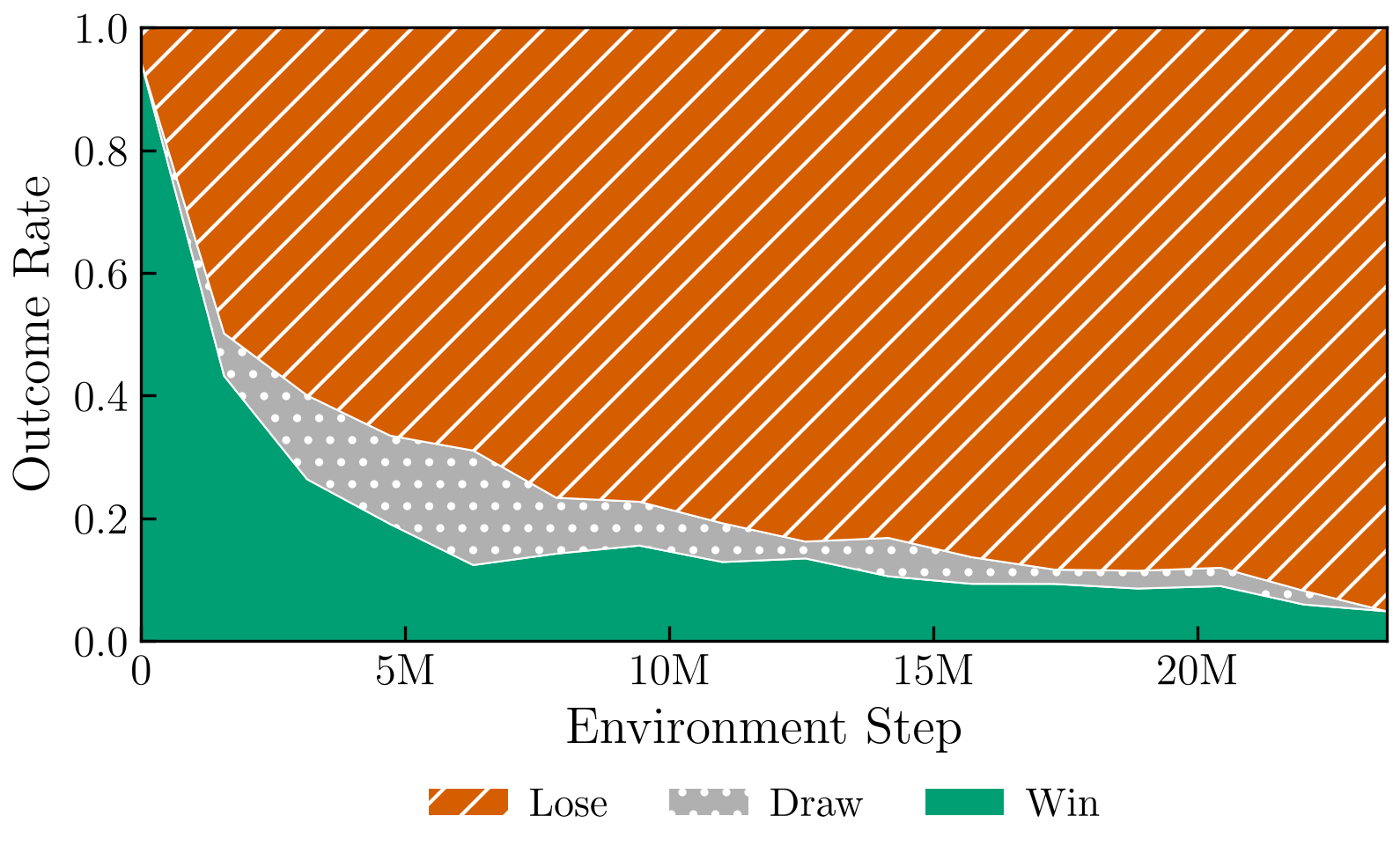}
    \caption{EMAgnet}
    \label{fig:emagnet_exploiter}
  \end{subfigure} 
  \begin{subfigure}[t]{0.48\linewidth}
    \centering
    \includegraphics[width=\linewidth]{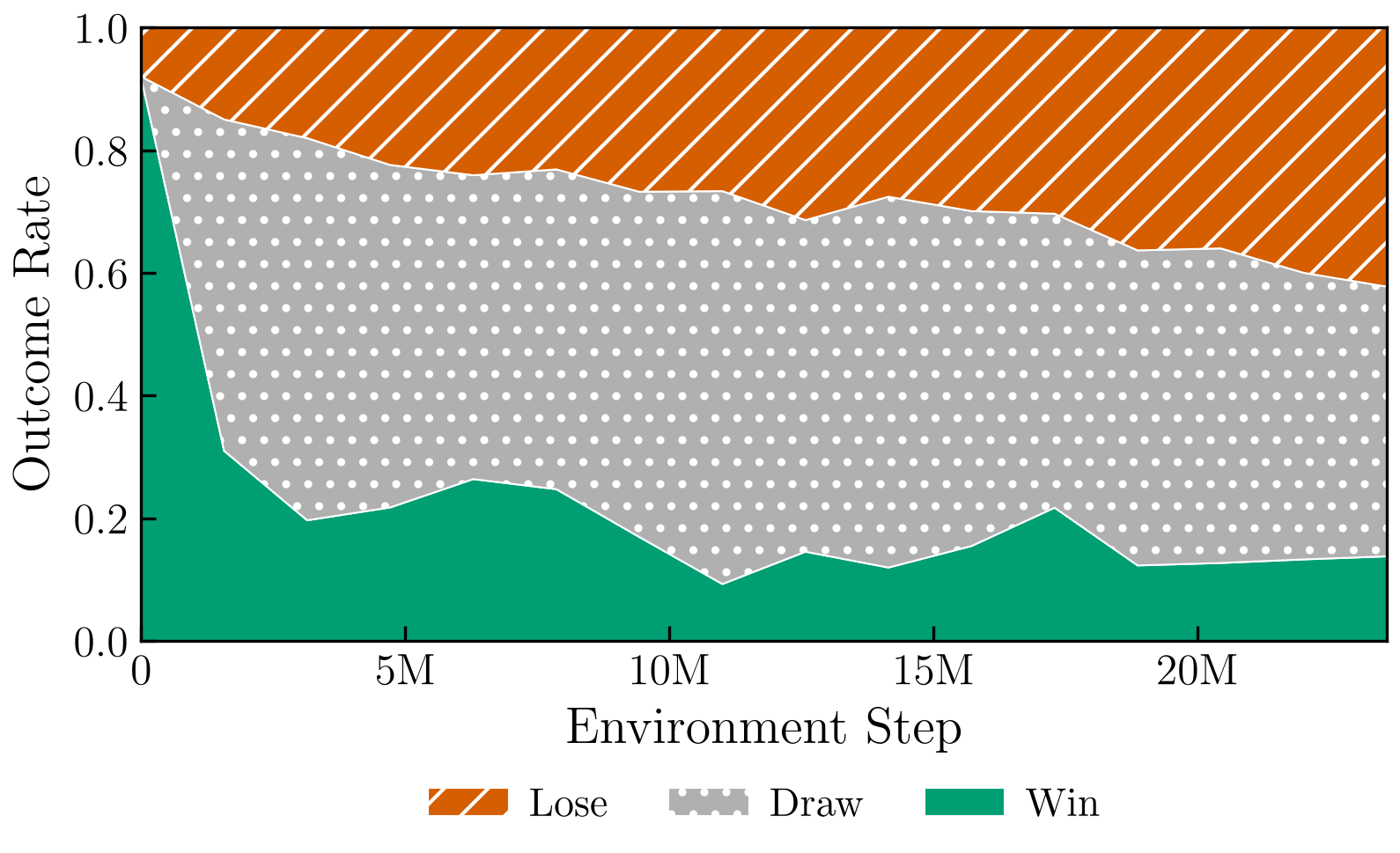}
    \caption{PFSP}
    \label{fig:pfsp_exploiter}
  \end{subfigure}
  \caption{Results of training approximate best responses for all (\subref{fig:ppo_exploiter})~PPO, (\subref{fig:ppo_sched_exploiter})~PPO (Sched.), (\subref{fig:emagnet_exploiter})~EMAgnet, and (\subref{fig:pfsp_exploiter})~PFSP policies. Results are presented from the perspective of the trained policy (i.e., the loss rate is how often the best response policies win). PPO and PPO (Sched.) perform approximately equivalently. EMAgnet fares the worst, with the highest loss rate and the fewest ties. PFSP has a relatively high tie rate. The win rate is uniformly low across all algorithms. Three random seeds are used for each best response, resulting in 15 total for each algorithm (3 best response seeds $\times$ 5 algorithm seeds).}
  \label{fig:approx_exploitability}
\end{figure}

\paragraph{Approximate Exploitability.}
While the heuristic opponents provide a coarse view of skill, they say little about how robust each policy is to a dedicated adversary. We therefore also report the approximate exploitability of each set of policies by training approximate best responses to each. We train best responses using PPO against the fixed final checkpoint of each policy. This approach has become a common approximation of exploitability in games where it is intractable to calculate exact exploitability~\citep[e.g.,][]{lanctot2019openspiel, timbers2020approximate, sokota2022unified}. The parameters used are identical to those of the fixed-entropy PPO policies (described in Appendix~\ref{sec:appendix_alg_details}). The results for each algorithm, in terms of the game outcome rates, are shown in Figure~\ref{fig:approx_exploitability}. We observe relatively consistent results, with statistically significant differences between EMAgnet and PFSP in terms of final win rate and draw rate, with the latter performing more strongly against the best response (Holm-corrected Mann--Whitney $U$ test, $p=0.048$ and $p=0.045$, respectively). There are no significant differences in any corrected pairwise comparisons with either PPO variant.

\begin{figure}[t]
  \centering
    \begin{subfigure}[t]{0.48\linewidth}
    \centering
    \includegraphics[width=\linewidth]{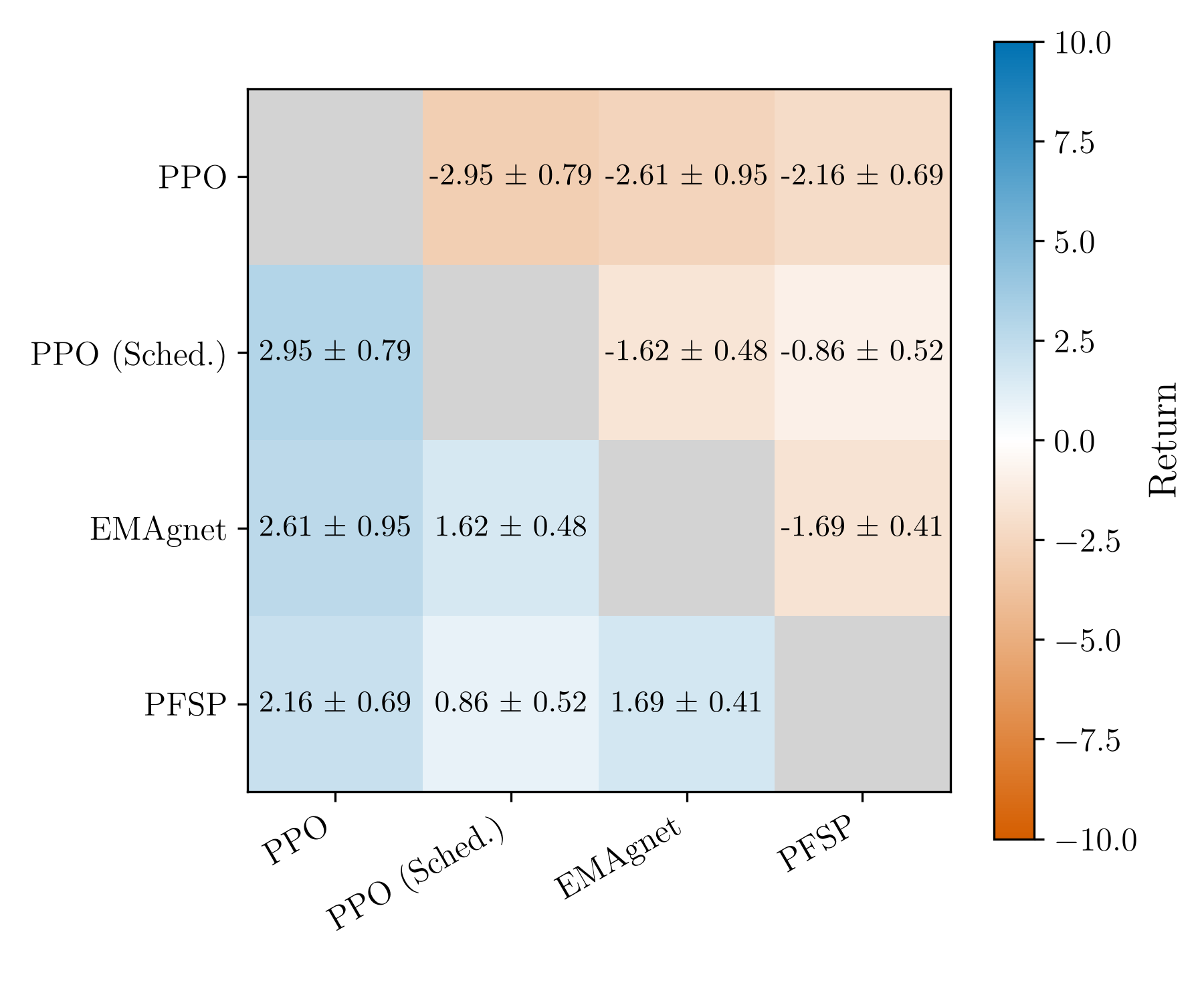}
    \caption{Head-to-Head Returns}
    \label{fig:head_to_head}
  \end{subfigure} 
  \begin{subfigure}[t]{0.48\linewidth}
    \centering
    \includegraphics[width=\linewidth]{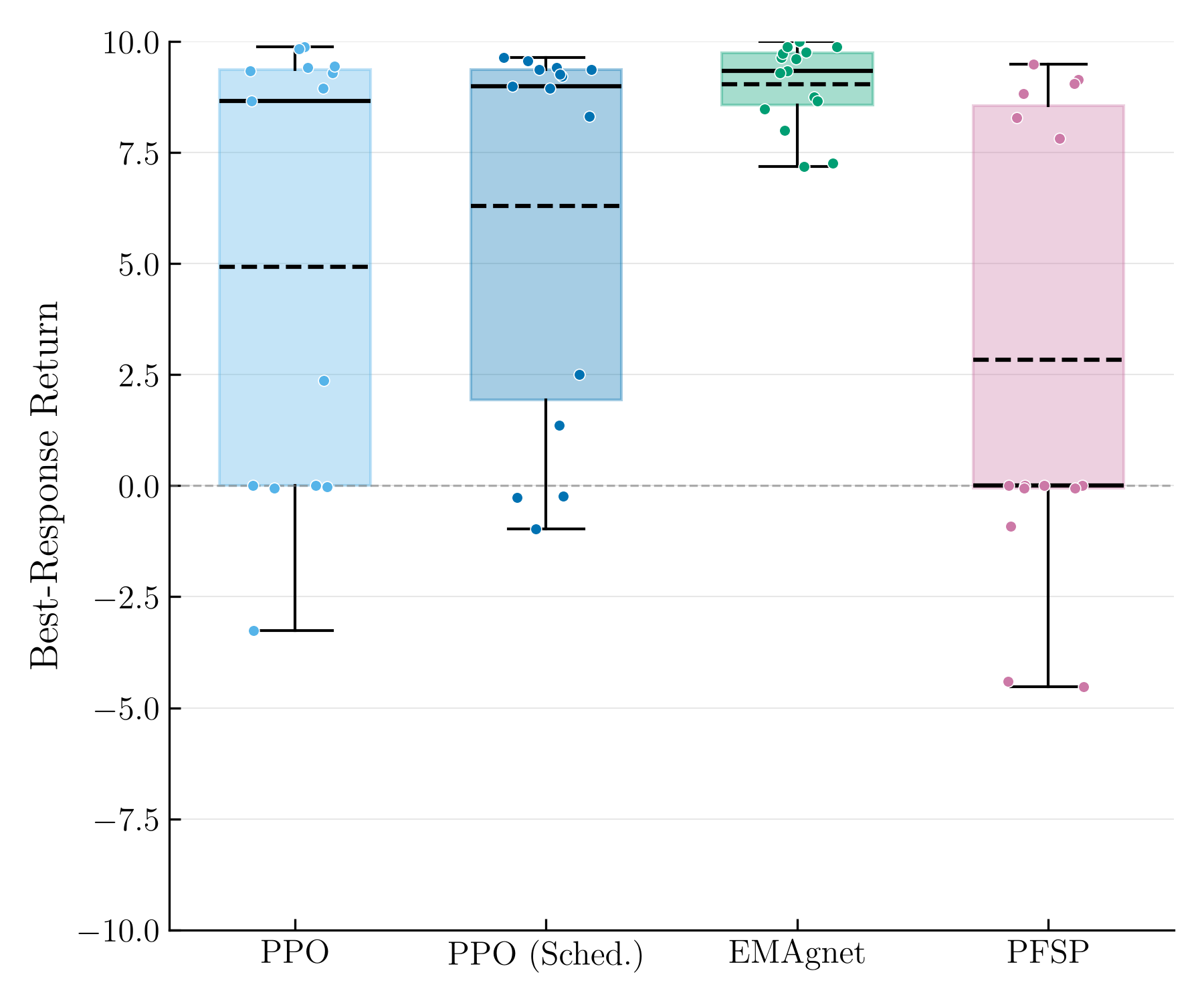}
    \caption{Best Response Returns}
    \label{fig:exploiter_returns}
  \end{subfigure} 
  
  \caption{(\subref{fig:head_to_head})~Head-to-head and (\subref{fig:exploiter_returns})~best response returns for all policies. PFSP has a positive return against all other algorithms. EMAgnet has a positive return against both PPO variants, but the best responses trained against it consistently attain the highest returns. In (\subref{fig:head_to_head}), each cell shows the expected return to the row policies when playing against the column policies (mean $\pm$ SEM). Each seed of the row algorithm plays each seed of the column algorithm for 96 games, for a total of $5\times 5\times 96=2{,}400$ games per cell. The return of the final best response checkpoints trained against each algorithm is represented in (\subref{fig:exploiter_returns}), with each point representing ${\sim}140$ games for each individual best response policy. Boxes span the interquartile range; the solid line is the median and the dashed line is the mean.}
  \label{fig:returns}
\end{figure}

To examine the relationship between skill and exploitability, Figure~\ref{fig:returns} shows both the head-to-head returns of all policies, as a proxy for skill, and the final returns of the best responses. PFSP has a positive return against all other policies, despite no statistically significant difference in exploitability relative to the PPO variants. Similarly, EMAgnet has a positive return against both PPO variants, but is more consistently exploited. 

\begin{figure}
    \centering
    \includegraphics[width=0.5\linewidth]{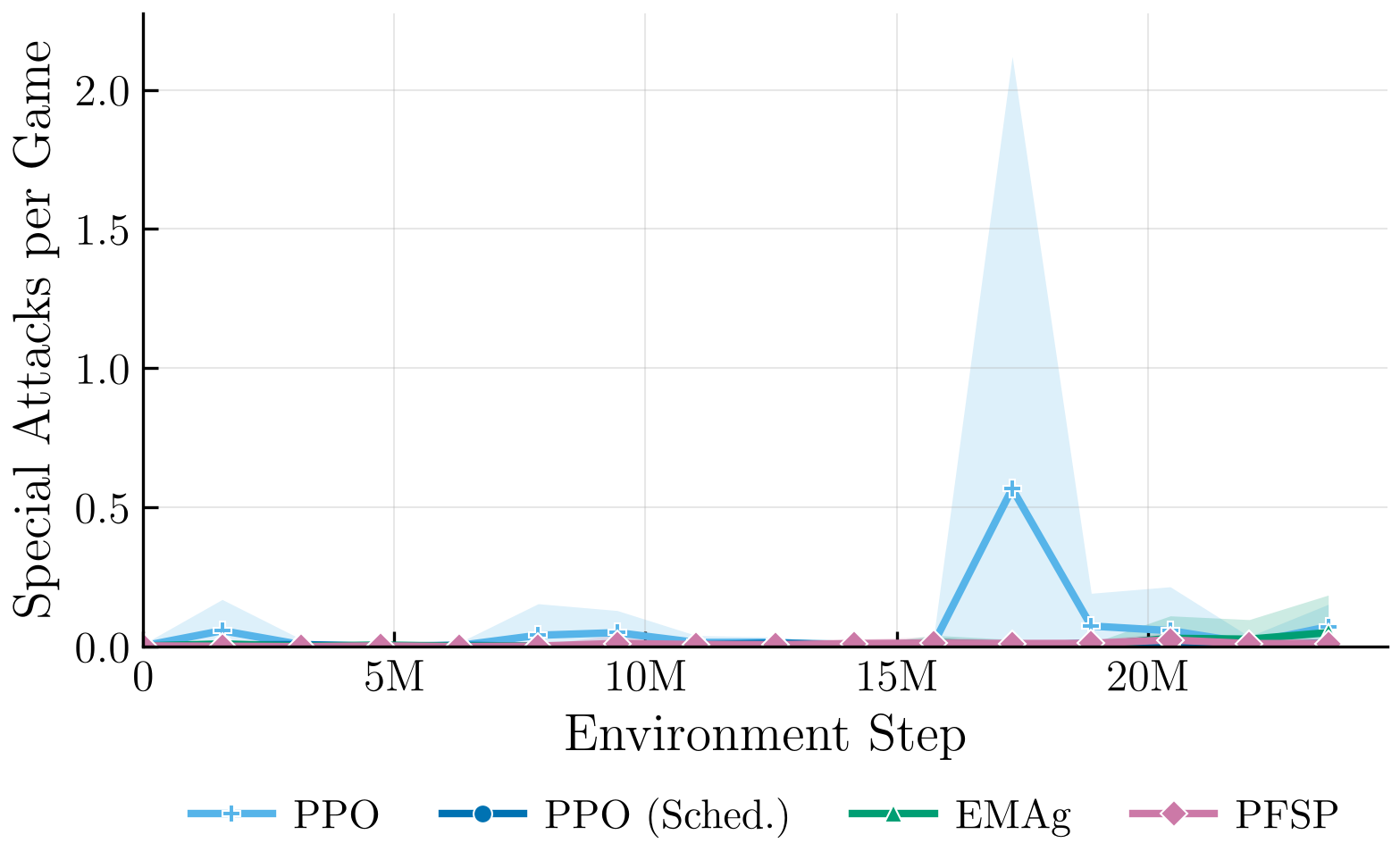}
    \caption{Usage of the directional special attack (\texttt{B\_SPECIAL}) across algorithms. In the standard environment configuration, the move requires selecting any attack action (\textsc{attack}, \textsc{forward+attack}, or \textsc{back+attack}) for 15 contiguous steps (60 frames), followed by either the \textsc{forward} or \textsc{back} action.}
    \label{fig:sp_charge_usage}
\end{figure}

\paragraph{Discovering Special Attacks.}
Lastly, we show a distinguishing feature of FootsiesGym relative to other small-scale benchmark environments. As discussed in Section~\ref{sec:env} (and further in Appendix~\ref{sec:appendix_sp_charge}), the special attack actions are difficult to discover and learn to use effectively through random exploration. In particular, the directional special attack (\texttt{B\_SPECIAL}) requires a player to select an attack action for 15 consecutive steps (at a frame skip of 4), releasing into either the \textsc{forward} or \textsc{backward} action. There is a roughly 0.001\% chance of this sequence occurring under a uniform random policy. The difficulty of learning this move is exacerbated by the fact that successful execution is dependent on the opponent's behavior. Figure~\ref{fig:sp_charge_usage} illustrates the usage of this move throughout training for all algorithms. It shows that throughout training, the average number of \texttt{B\_SPECIAL} executions is near zero, with the one exception of the fixed-entropy PPO policy learning it for a brief period before it leaves the strategy space. This illustrates the complexity of the environment, as well as a potential incompatibility with standard game-theoretic training methods. Indeed, increasing entropy regularization, as is effective in small-scale benchmarks, may make it increasingly difficult for a policy to learn to utilize this move effectively. FootsiesGym also includes an option to augment the action space to make the use of special moves easier, as illustrated in Appendix~\ref{sec:appendix_sp_charge}.

\section{Discussion and Future Work} \label{sec:discussion}

% =====================================================================
% Page budget: ~0.5 page.
% Three forward-looking paragraphs + a 1-2 sentence limitation note.
% =====================================================================

\paragraph{Game Theoretic Training.} As previously discussed, FootsiesGym offers meaningful strategic complexity while remaining resource-efficient. This makes it a natural testbed for comparing, validating, and developing game-theoretic training regimes such as population-based training~\citep{vinyals2019grandmaster}, PSRO~\citep{lanctot2017unified}, or regularization-based methods~\citep{perolat2022mastering, sokota2022unified, rudolph2025reevaluating, maidment2026emagnet}. Our results also surface a concrete challenge for such methods: strategies that are valuable but hard to discover, such as the charged special attacks, may be driven out of the strategy space by the same regularization that these approaches rely on (Section~\ref{sec:experiments}). Understanding how game-theoretic training can retain such strategies is an open question that FootsiesGym is well positioned to study.

\paragraph{Beyond Skill.}
Beyond raw skill or exploitability minimization, an open question is what makes a policy \emph{feel} like a compelling opponent rather than merely a strong one. In Section~\ref{sec:experiments}, we demonstrated that all baseline policies increase in skill but become increasingly reactive over time, failing to maintain a high win rate against a stationary opponent. We also showed that the learned policies almost never discover or consistently use special attacks, despite these being a core mechanic of the game. While high-skill policies are desirable in some settings, others may instead value policies that proactively engage opponents and make effective use of all core game mechanics, including special attacks. A growing body of work pursues this direction: \citet{zhao2020winning} argue that ``winning isn't everything,'' surveying agents designed for human-like behavior, playtesting, and game balancing rather than pure competitive optimality. Our environment illustrates a failure case for engaging play and prompts the question of how to develop algorithms that meet both game-theoretic and play-style requirements.

\paragraph{Exploitability at Scale.}
In game-theoretic environments, exploitability quickly becomes intractable to calculate directly. Indeed, sophisticated approaches are required for even relatively simple games~\citep{rudolph2025reevaluating}, while more complex games require rough approximations, such as directly training a best response~\citep{sokota2022unified}. FootsiesGym, while being too complex for exact calculation, may provide a fruitful testbed for identifying approximate methods due to its speed and balanced complexity. Our baselines illustrate why such measures matter: EMAgnet attains positive head-to-head returns against both PPO variants yet is the most consistently exploited (Section~\ref{sec:experiments}), a disconnect that win-rate comparisons alone would not reveal.

\paragraph{Human Evaluation and Imitation Learning.}
Through the Unity game engine, the Footsies game can be exported to a WebGL build and deployed in the browser for human-subject experiments, enabling human-vs-AI evaluation and data collection for imitation learning. Such data would also ground the play-style questions raised above: evaluation versus human participants affords the ability to directly assess whether a policy's behavior is perceived as engaging or fun.

\paragraph{Limitations.}
Footsies is a complex, real-time, imperfect-information game relative
to many RL benchmarks. Nevertheless, there remains a substantial complexity gap
between it and modern commercial fighting games. In addition, our reported
exploitability is estimated with approximate best responses, which only lower-bound
true exploitability~\citep{timbers2020approximate}, and our baselines are only
na\"{\i}vely tuned; both caveats should be kept in mind when interpreting the
comparisons in Section~\ref{sec:experiments}.
%%%%%%%%%%%%%%%%%%%%%%%%%%%%%%%%%%%%%%%%%%%%%%%%%%%%%%%%%%%%%%%%
%% Appendices
%%%%%%%%%%%%%%%%%%%%%%%%%%%%%%%%%%%%%%%%%%%%%%%%%%%%%%%%%%%%%%%%
\appendix

\section{Environment configuration keys} \label{sec:appendix_config}

Table~\ref{tab:env_config} displays the environment configurations that are accessible to users. In all training runs, we use the default environment parameters. The exceptions are our demonstrations of the effects of manipulating \texttt{action\_delay} and \texttt{use\_special\_charge\_action}, which are illustrated in Appendix~\ref{sec:appendix_delay} and Appendix~\ref{sec:appendix_sp_charge}, respectively.

\begin{table}[h]
  \centering
  \small
  \setlength{\tabcolsep}{6pt}
  \begin{tabular}{lll}
    \toprule
    Key & Default & Description \\
    \midrule
    \texttt{num\_envs}                     & $1$       & Parallel environments per process \\
    \texttt{frame\_skip}                   & $4$       & Game frames per environment step (60\,Hz native) \\
    \texttt{action\_delay}                 & $8$       & Number of frames between action selection and execution \\
    \texttt{max\_t}                        & $4000$    & Truncation horizon (environment steps) \\
    \texttt{use\_special\_charge\_action}  & \texttt{False} & Expand action space to include charge toggles \\
    \texttt{win\_reward\_scaling\_coeff}   & $10.0$     & Magnitude of the terminal win/loss reward \\
    \texttt{guard\_break\_reward}          & $0.0$     & Dense per-guard-break shaped reward \\
    \texttt{use\_reward\_budget}           & \texttt{False} & Deduct shaping from the win-reward budget \\
    \bottomrule
  \end{tabular}
  \caption{Selected FootsiesEnv config keys. \texttt{action\_delay} must
    be a multiple of \texttt{frame\_skip}.}
  \label{tab:env_config}
\end{table}

\section{Observation Features} \label{sec:appendix_obs}

Table~\ref{tab:obs_features} lists the features that
make up each agent's observation. The Common block (distance between
players) appears once. The Shared block appears twice, once for self
and once for opponent. The Privileged block appears only for self,
modeling the information asymmetry between two humans seated at
separate controllers (e.g., a player does not directly see their
opponent's special-attack charge timer). In total, the observation comprises $85$ dimensions with the default action space and $88$ dimensions with the special-charge option enabled.
\begin{table}[h]
  \centering
  \small
  \setlength{\tabcolsep}{6pt}
  \begin{tabular}{llcl}
    \toprule
    Block & Feature & Dim & Encoding \\
    \midrule
    Common      & inter-player distance ($|x_{p_1} - x_{p_2}|$) & 1 & scalar, normalized by stage width \\
    \midrule
    Shared      & \texttt{player\_position\_x}            & 1 & scalar, normalized \\
                & \texttt{velocity\_x}                    & 1 & scalar, normalized \\
                & \texttt{is\_dead}                       & 1 & binary \\
                & \texttt{vital\_health}                  & 1 & scalar \\
                & \texttt{guard\_health}                  & 4 & one-hot over $\{0,1,2,3\}$ \\
                & \texttt{current\_action\_id}            & 17 & one-hot over player script IDs \\
                & \texttt{current\_action\_frame}         & 1 & scalar, normalized \\
                & \texttt{current\_action\_frame\_count}  & 1 & scalar, normalized \\
                & \texttt{current\_action\_remaining}     & 1 & scalar, normalized \\
                & \texttt{is\_action\_end}                & 1 & binary \\
                & \texttt{is\_always\_cancelable}         & 1 & binary \\
                & \texttt{current\_action\_hit\_count}    & 1 & scalar \\
                & \texttt{current\_hit\_stun\_frame}      & 1 & scalar, normalized \\
                & \texttt{is\_in\_hit\_stun}              & 1 & binary \\
                & \texttt{sprite\_shake\_position}        & 1 & scalar \\
                & \texttt{max\_sprite\_shake\_frame}      & 1 & scalar, normalized \\
                & \texttt{is\_face\_right}                & 1 & binary \\
                & \texttt{current\_frame\_advantage}      & 1 & scalar, normalized \\
    \midrule
    Privileged  & \texttt{would\_next\_forward\_input\_dash}  & 1 & binary \\
                & \texttt{would\_next\_backward\_input\_dash} & 1 & binary \\
                & \texttt{special\_attack\_progress}          & 1 & scalar, clipped to $[0,1]$ \\
                & \texttt{previous\_action}                   & $|\mathcal{A}|$ & one-hot over agent actions \\
                & \texttt{is\_holding\_special\_charge}       & 1 & binary \\
    \bottomrule
  \end{tabular}
  \caption{Per-feature breakdown of each agent's observation. Each
    observation contains the Common block (distance), two copies of
    the Shared block (self and opponent), and one copy of the
    Privileged block (self only). $|\mathcal{A}|$ is the size of the
    agent action space (6 by default, 9 with the special-charge
    action option enabled).}
  \label{tab:obs_features}
\end{table}

\section{Algorithm Details} \label{sec:appendix_alg_details}

In the following sections, we detail the parameters used for each of the three learning algorithms. 

\subsection{Proximal Policy Optimization (PPO)} \label{sec:ppo_hparams}

Table~\ref{tab:ppo_hparams} lists the algorithm, optimization, and
architecture hyperparameters used for both PPO self-play baselines in Section~\ref{sec:experiments}. The two variants differ only in the entropy coefficient: PPO (Sched.) linearly anneals it from $0.1$ to $0$, while PPO holds it fixed at $0.025$.

\begin{table}[h]
  \centering
  \small
  \setlength{\tabcolsep}{6pt}
  \begin{tabular}{ll}
    \toprule
    Hyperparameter & Value \\
    \midrule
    Total timesteps                     & $2.5\times 10^{7}$    \\
    Parallel game instances             & $48$                \\
    Rollout length                      & $64$                \\
    Update epochs                       & $8$                 \\
    Minibatches per epoch               & $8$                 \\
    Discount $\gamma$                   & $0.995$             \\
    GAE $\lambda$                       & $0.95$              \\
    PPO clip $\epsilon$                 & $0.3$               \\
    Value-loss coefficient              & $1.0$               \\
    Entropy coefficient                 & $0.1 \to 0$ (Sched.); $0.025$ (fixed)         \\
    Max gradient norm                   & $0.5$               \\
    Learning rate                       & $1\times 10^{-4} \to 0$ \\
    Optimizer                           & Adam                \\
    Architecture                        & MLP, 2 hidden layers of $256$ + ReLU \\
    \bottomrule
  \end{tabular}
  \caption{PPO hyperparameters used for the baselines in
    Section~\ref{sec:experiments}. The learning rate and the PPO (Sched.)
    entropy coefficient follow linear schedules.}
  \label{tab:ppo_hparams}
\end{table}

\subsection{EMAgnet} \label{sec:appendix_emagnet}

EMAgnet was introduced by~\citet{maidment2026emagnet} and represents an alternative regularization scheme for policy gradient methods. We apply it to PPO, augmenting the loss function with a KL divergence term towards a policy with the exponential moving average of the weights, described in Equation~\ref{eqn:emag_loss}.

\begin{equation} \label{eqn:emag_loss}
  \mathcal{L}_{\text{PPO-EMAg}}(\theta)
  = \mathcal{L}_{\text{PPO}}(\theta)
  + \lambda_{\mathrm{KL}} \, \mathbb{E}_{z \sim \mathcal{T}}\!\bigl[D_{\mathrm{KL}}\bigl(\pi_{\theta_{\mathrm{mag}}}(\cdot \mid z) \,\|\, \pi_\theta(\cdot \mid z)\bigr)\bigr],
\end{equation}
where $\theta_{\mathrm{mag}}$ denotes the magnet parameters and $\lambda_{\mathrm{KL}} > 0$ controls the regularization strength. After each PPO update, the magnet parameters are updated as
\begin{equation}
  \theta_{\mathrm{mag}} \leftarrow (1 - \tau) \, \theta_{\mathrm{mag}} + \tau \, \theta,
\end{equation}
with step size $\tau \in (0,1]$. The parameters we used are provided in Table~\ref{tab:emag_hparams}.

\begin{table}[h]
  \centering
  \small
  \setlength{\tabcolsep}{6pt}
  \begin{tabular}{ll}
    \toprule
    Hyperparameter & Value \\
    \midrule
    Total timesteps                     & $2.5\times 10^{7}$    \\
    Parallel game instances             & $48$                \\
    Rollout length                      & $64$                \\
    Update epochs                       & $8$                 \\
    Minibatches per epoch               & $8$                 \\
    Discount $\gamma$                   & $0.995$             \\
    GAE $\lambda$                       & $0.95$              \\
    PPO clip $\epsilon$                 & $0.3$               \\
    Value-loss coefficient              & $1.0$               \\
    Entropy coefficient                 & $3\times 10^{-3}$         \\
    Max gradient norm                   & $0.5$               \\
    Learning rate                       & $1\times 10^{-3} \to 0$     \\
    Magnet update rate $\tau$           & $1\times 10^{-4}$              \\
    Magnet KL coefficient $\lambda_{\mathrm{KL}}$              & $0.5$              \\
    Optimizer                           & Adam                \\
    Architecture                        & MLP, 2 hidden layers of $256$ + ReLU \\
    \bottomrule
  \end{tabular}
  \caption{EMAgnet hyperparameters used for the baseline in
    Section~\ref{sec:experiments}. Learning rate utilizes a linear schedule.} 
  \label{tab:emag_hparams}
\end{table}

\subsection{Prioritized Fictitious Self-Play (PFSP)} \label{sec:appendix_pfsp}

We train PFSP agents using the prioritized fictitious self-play opponent-sampling
rule of \citet{vinyals2019grandmaster}, applied to a bounded reservoir of the
training agent's own past checkpoints rather than a full multi-agent league.
At fixed snapshot intervals, we snapshot the current policy into a
reservoir of $N$ slots; once the reservoir is full, new snapshots replace
existing ones via reservoir sampling. During training, each game is played
against the current policy with probability $p_\mathrm{sp}=0.8$ (mirror
self-play) and against a stored snapshot otherwise. The remaining
$1-p_\mathrm{sp}$ of the opponent budget is distributed over the snapshots in
proportion to a priority that favors opponents the agent is currently losing to.

Following the weighting of \citet{vinyals2019grandmaster}, snapshot
$i$ is sampled with weight
\begin{equation}
  w_i \;\propto\; \bigl(1 - \hat{p}_i\bigr)^{1/2},
\end{equation}
where $\hat{p}_i$ is the estimated win rate of the current policy against snapshot
$i$; snapshots the agent loses to more often are thus played more frequently. The
win rates $\hat{p}_i$ are refreshed on the same cadence as snapshotting through
dedicated evaluation games and smoothed with an exponential moving average, and a
slot is reset to a neutral $\hat{p}_i=0.5$ whenever it is overwritten. Unlike the original formulation, we construct the opponent pool from checkpoints of a single continuously trained policy rather than maintaining a league of independently trained best-response agents. The complete hyperparameters are listed in
Table~\ref{tab:pfsp_hparams}.

\begin{table}[h]
  \centering
  \small
  \setlength{\tabcolsep}{6pt}
  \begin{tabular}{ll}
    \toprule
    Hyperparameter & Value \\
    \midrule
    Total timesteps                      & $2.5\times 10^{7}$ \\
    Parallel game instances              & $48$ \\
    Rollout length                       & $64$ \\
    Update epochs                        & $8$ \\
    Minibatches per epoch                & $8$ \\
    Discount $\gamma$                    & $0.995$ \\
    GAE $\lambda$                        & $0.95$ \\
    PPO clip $\epsilon$                  & $0.3$ \\
    Value-loss coefficient               & $1.0$ \\
    Entropy coefficient                  & $0.1 \to 0$ \\
    Max gradient norm                    & $0.5$ \\
    Learning rate                        & $1\times 10^{-4} \to 0$ \\
    Reservoir size $N$                   & $16$ \\
    Snapshot interval (env.\ steps)      & $614{,}400$ \\
    Self-play proportion $p_\mathrm{sp}$ & $0.8$ \\
    Optimizer                            & Adam \\
    Architecture                         & MLP, 2 hidden layers of $256$ + ReLU \\
    \bottomrule
  \end{tabular}
  \caption{PFSP hyperparameters used for the baseline in
    Section~\ref{sec:experiments}. Learning rate and entropy coefficient use linear schedules.}
  \label{tab:pfsp_hparams}
\end{table}

\section{Effect of Action Delay on Policy Quality} \label{sec:appendix_delay}

\begin{figure}[h]
  \centering
  \begin{subfigure}[t]{0.48\linewidth}
    \centering
    \includegraphics[width=\linewidth]{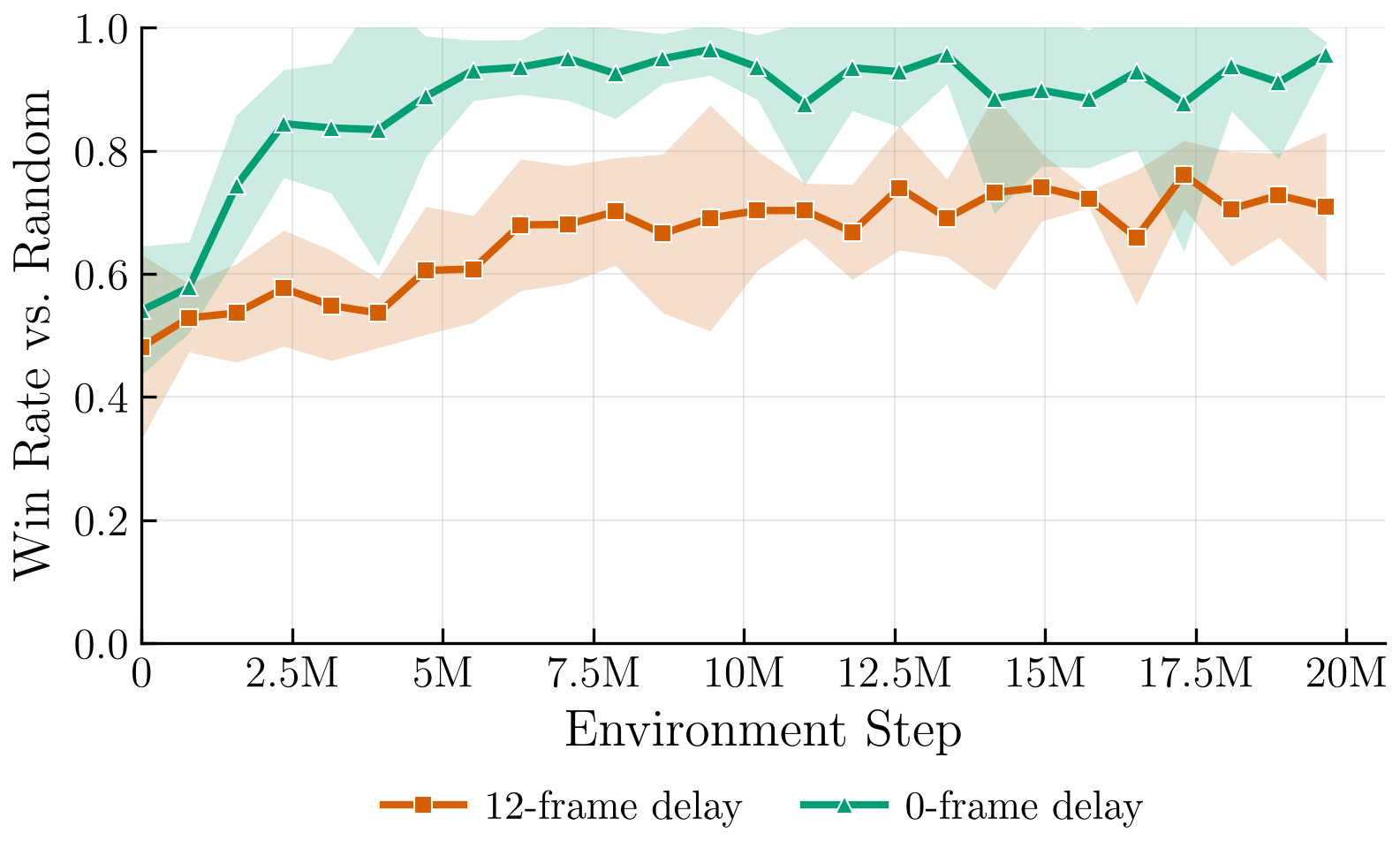}
    \caption{vs.\ random opponent.}
    \label{fig:ppo_curve_delay_random}
  \end{subfigure}
  \hfill
  \begin{subfigure}[t]{0.48\linewidth}
    \centering
    \includegraphics[width=\linewidth]{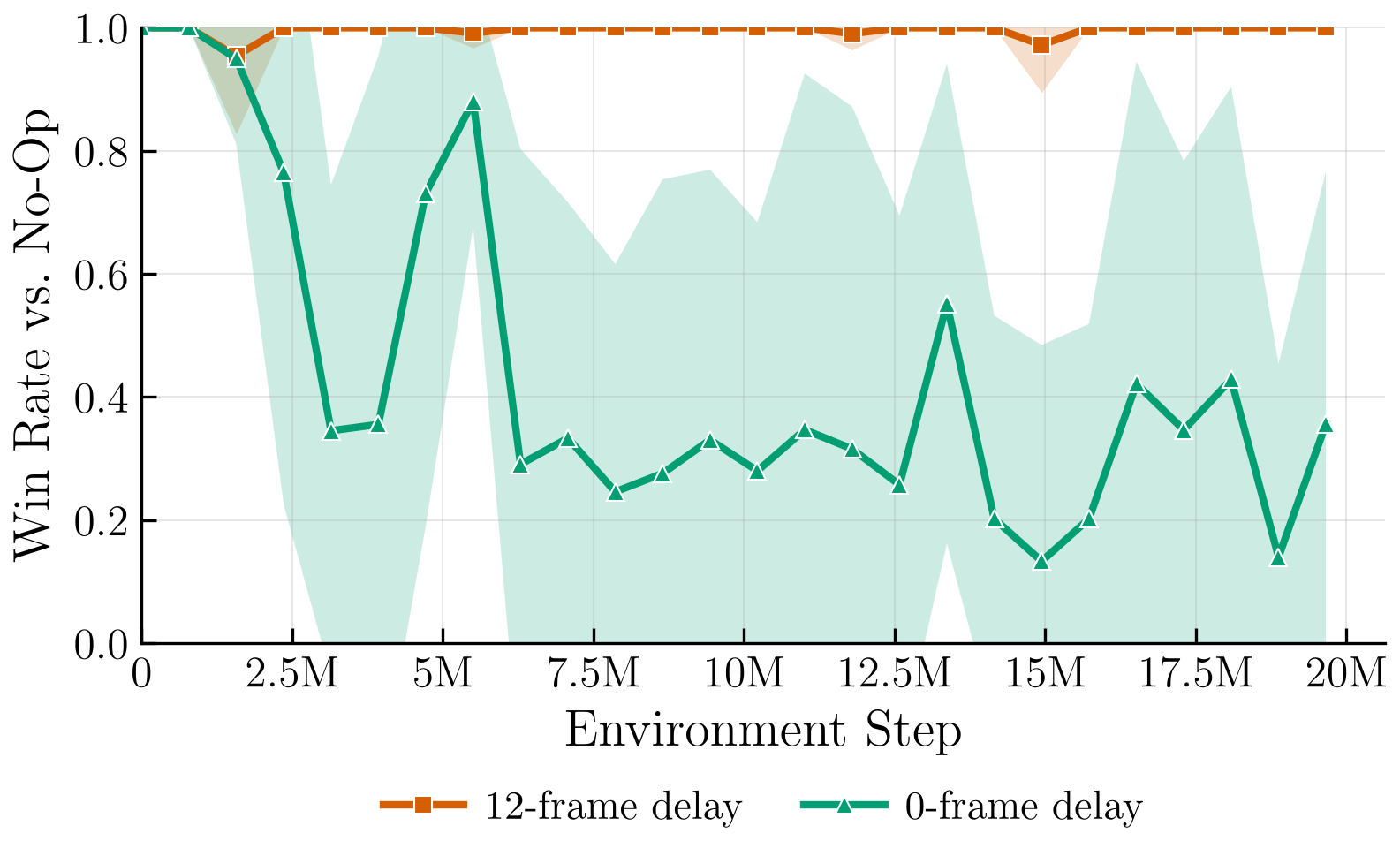}
    \caption{vs.\ no-op opponent.}
    \label{fig:ppo_curve_delay_noop}
  \end{subfigure}
  \caption{PPO self-play policies trained with
    \texttt{action\_delay}\,$\in\{0, 12\}$, evaluated against a
    random opponent (\subref{fig:ppo_curve_delay_random}) and a no-op
    opponent (\subref{fig:ppo_curve_delay_noop}). 
    Curves are aggregated over 5 seeds; the line is the mean and the
    shaded band is the 95\% confidence interval.}
  \label{fig:ppo_curve_delay}
\end{figure}

Figure~\ref{fig:ppo_curve_delay} compares PPO self-play policies
trained with action delays of 0 and 12. All other
hyperparameters and environment settings are identical to those in
Appendix~\ref{sec:ppo_hparams}. Although the 0-delay policy attains
a higher win rate against the random opponent, it is completely
degenerate against the no-op opponent. Consistent with visual
inspection, the 0-delay policy is brittle in general despite its
strong numerical performance against random. On the other hand, the 12-delay policy has worse performance against random, but has a near 100\% win rate against no-op. 

The impact of action delay manipulations is inherently tied to the frame data in Footsies. Whether a move is reactable or must be predicted is determined by the length of its startup, active, and recovery frames. Manipulations of the action delay lever (and frame skip) should be informed by the game's frame data. The frame data of the attacks is shown in Table~\ref{tab:frame_data}.

\begin{table}[h]
  \centering
  \small
  \begin{tabular}{lccc}
    \toprule
    Attack & Startup & Active & Recovery \\
    \midrule
    \texttt{N\_ATTACK} & 4  & 2 & 16 \\
    \texttt{B\_ATTACK} & 3  & 3 & 15 \\
    \texttt{N\_SPECIAL} & 11 & 4 & 29 \\
    \texttt{B\_SPECIAL} & 2  & 6 & 47 \\
    \bottomrule
  \end{tabular}
  \caption{Frame data for the four attacks in Footsies. Startup is the window to block on reaction; recovery is the window to
    whiff-punish on reaction. Whiff-punishing refers to attacking a player that missed an attack.}
  \label{tab:frame_data}
\end{table}

\begin{figure*}[t]
  \centering
  \includegraphics[width=\textwidth]{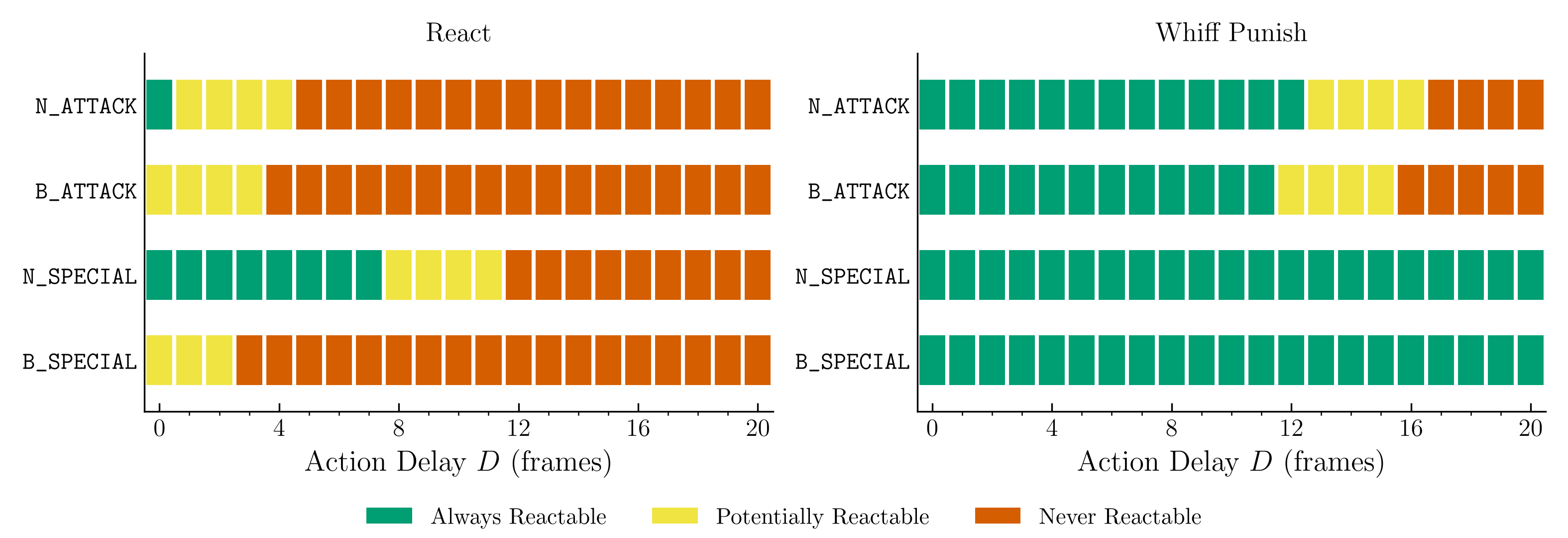}
  \caption{Reactability of each attack across action delays
    $D$. A stimulus can land anywhere in the skip window,
    adding $0$ frames in the best case and the full frame skip (4 in this figure) in the worst case. As a result, each delay is
    one of three states: Always Reactable, Potentially Reactable, or Never
    Reactable. The reaction window is dictated by the attack's startup or recovery frames. 
    At zero delay every attack is at least potentially reactable, but only \texttt{N\_ATTACK} and
    \texttt{N\_SPECIAL} are reactable in the worst case; both special attacks have long recoveries that permit
    punishment at any reasonable delay.}
  \label{fig:reactability}
\end{figure*}

The frame data allows us to calculate whether or not an attack is reactable at a given action delay. We illustrate this in Figure~\ref{fig:reactability}.

\section{Effect of Enabling Special Charge Actions} \label{sec:appendix_sp_charge}

There are two distinct special attacks in Footsies (scripts \texttt{N\_SPECIAL} and \texttt{B\_SPECIAL}). Both, if landed on a vulnerable opponent, win the round. The special attacks can be executed in several different ways. First, both can be executed by ``charging'' the attack action. If the attack action is held for at least 60 frames (or 15 steps at a frame skip of 4) then released, a special attack executes. If either the forward or backward directional actions are pressed when attack is released, the \texttt{B\_SPECIAL} move is executed. If no directional modification is applied when the attack is released, \texttt{N\_SPECIAL} is executed. Further, \texttt{N\_SPECIAL} can also be triggered in a combination attack: if a regular attack is already being executed and the attack action is selected again, a combination is triggered that executes \texttt{N\_SPECIAL} as a follow-up to the regular attack.

As discussed in Section~\ref{sec:experiments}, discovering the effective use of special attacks is difficult under random exploration due to the extended ``charge'' duration and strategic timing requirement. The \texttt{use\_special\_charge\_action} option in FootsiesGym is designed to alleviate this difficulty by adding additional actions that toggle the charge state of the special attack actions. All three of the added actions enable a charge state when first pressed: any action that is executed afterwards will result in the composite action with \textsc{attack} added to it (for example, \textsc{forward} becomes \textsc{forward+attack}). When pressed again, the charge state is disabled and a corresponding action is executed: \textsc{forward-charge} becomes \textsc{forward}, \textsc{back-charge} becomes \textsc{back}, and \textsc{neutral-charge} becomes \textsc{no-op}.

\begin{figure}[h]
  \centering
  \begin{subfigure}[t]{0.48\linewidth}
    \centering
    \includegraphics[width=\linewidth]{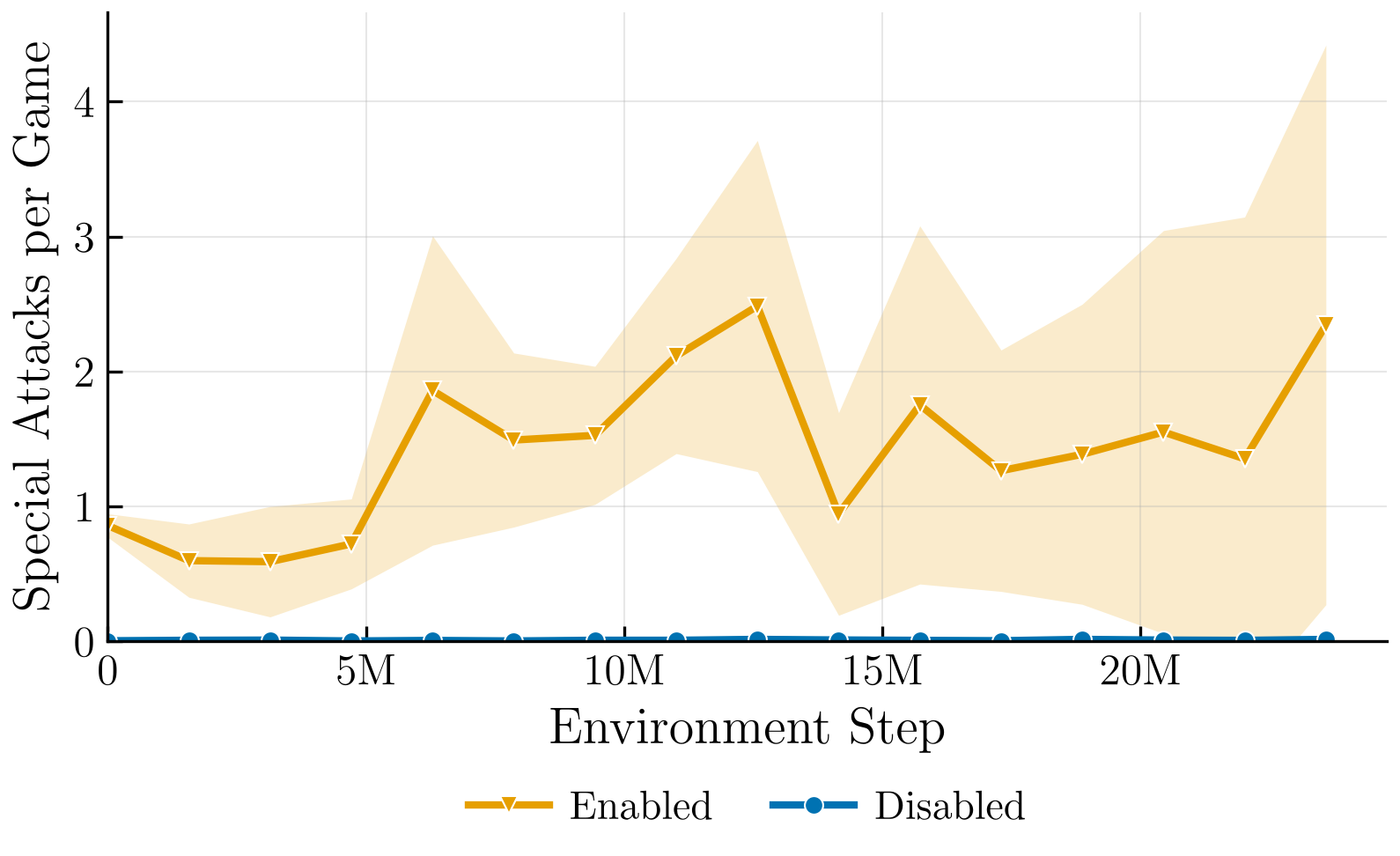}
    \caption{Usage of \texttt{B\_SPECIAL} action.}
    \label{fig:bspec}
  \end{subfigure}
  \hfill
  \begin{subfigure}[t]{0.48\linewidth}
    \centering
    \includegraphics[width=\linewidth]{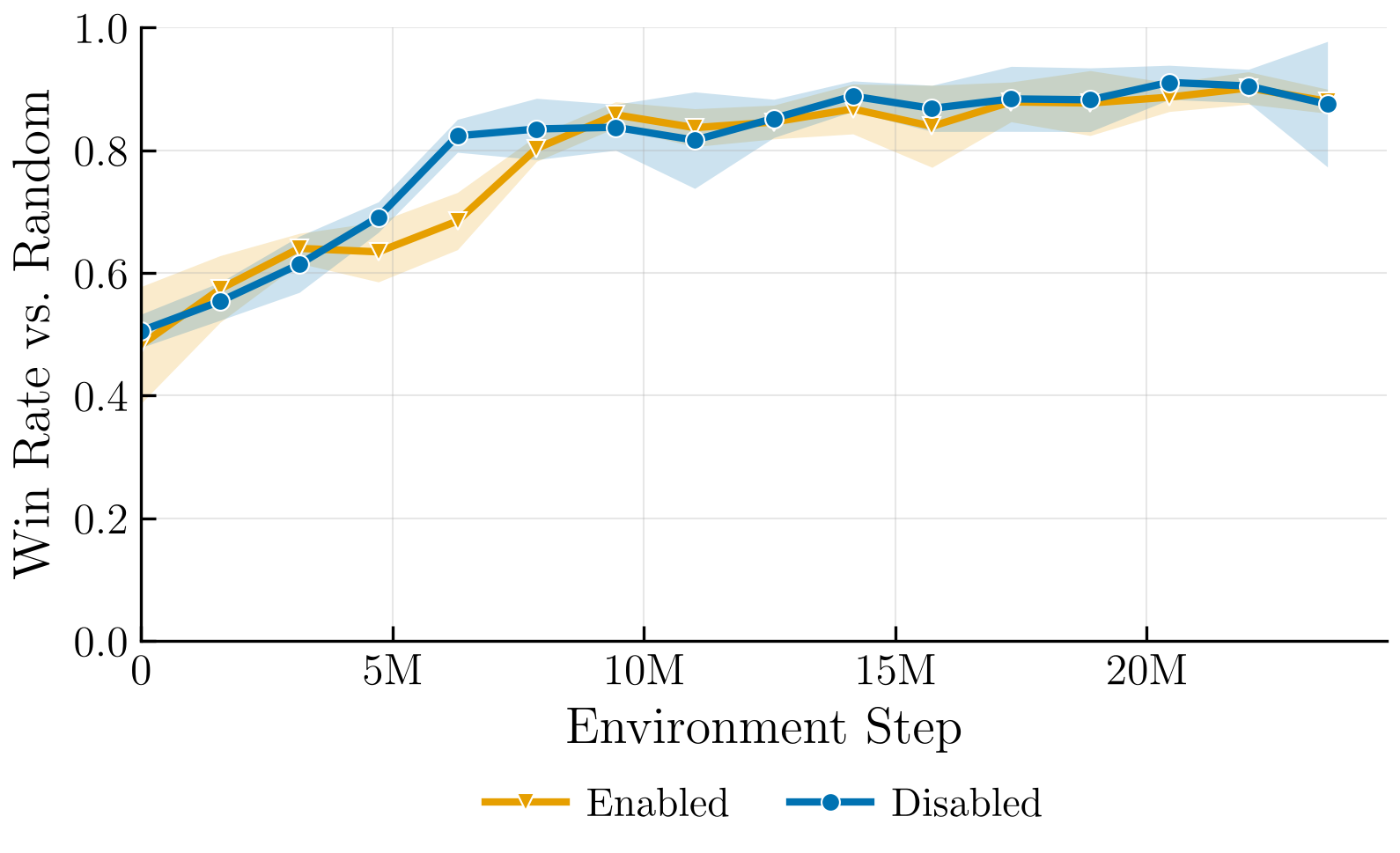}
    \caption{Win rate vs.\ random opponent with and without special charge actions.}
    \label{fig:sp_charge_vs_random}
  \end{subfigure}
  \caption{Comparison of PPO (Sched.) training with and without special charge actions enabled. The usage of \texttt{B\_SPECIAL} is shown in (\subref{fig:bspec}), and the win rate is shown in (\subref{fig:sp_charge_vs_random}).}
  \label{fig:spcharge_comparison}
\end{figure}

To illustrate the effect of this option, we train PPO (Sched.) with and without it enabled. Figure~\ref{fig:spcharge_comparison} shows move usage rates over training and performance against a random opponent. We see a substantial increase in usage of the move, along with comparable performance against the uniform random opponent.

% \label{sec:ack}
% Use unnumbered third level headings for the acknowledgments. All acknowledgments, including those to funding agencies, go at the end of the paper. Only add this information once your submission is accepted and deanonymized. The acknowledgments do not count towards the 8--12 page limit.

%%%%%%%%%%%%%%%%%%%%%%%%%%%%%%%%%%%%%%%%%%%%%%%%%%%%%%%%%%%%%%%%
%% NOTE: THIS MARKS THE END OF THE "MAIN TEXT"
%%%%%%%%%%%%%%%%%%%%%%%%%%%%%%%%%%%%%%%%%%%%%%%%%%%%%%%%%%%%%%%%

%%%%%%%%%%%%%%%%%%%%%%%%%%%%%%%%%%%%%%%%%%%%%%%%%%%%%%%%%%%%%%%%
%% Bibliography
%%%%%%%%%%%%%%%%%%%%%%%%%%%%%%%%%%%%%%%%%%%%%%%%%%%%%%%%%%%%%%%%
\bibliography{main}
\bibliographystyle{rlvg}

%%%%%%%%%%%%%%%%%%%%%%%%%%%%%%%%%%%%%%%%%%%%%%%%%%%%%%%%%%%%%%%%
% AUTHOR: If your paper has no supplementary materials, you may 
%         comment out the line below, which creates the title for
%         the supplementary materials.
%%%%%%%%%%%%%%%%%%%%%%%%%%%%%%%%%%%%%%%%%%%%%%%%%%%%%%%%%%%%%%%%
% \beginSupplementaryMaterials

% Content that appears after the references are not part of the ``main text,'' have no page limits, are not necessarily reviewed, and should not contain any claims or material central to the paper. 
% %
% If your paper includes supplementary materials, use the \begin{center}
%     {\tt {\textbackslash}beginSupplementaryMaterials} 
% \end{center}
% command as in this example, which produces the title and disclaimer above. 
% %
% If your paper does not include supplementary materials, this command can be removed or commented out.

\end{document}